\documentclass[letterpaper, 10 pt, conference]{ieeeconf}  

\IEEEoverridecommandlockouts                              

\usepackage{times}
\usepackage{amsmath,amssymb,amsfonts}
\usepackage{optidef} 
\usepackage{bm}
\usepackage{acronym}
\usepackage{graphicx}
\usepackage{algorithm2e}
\usepackage{caption}
\usepackage{comment}
\usepackage{xcolor}
\usepackage{booktabs}
\usepackage{url}
\usepackage[nocompress]{cite}

\usepackage[switch]{lineno}

\definecolor{ourblue}{rgb}{0.368,0.507,0.71}
\definecolor{ourgreen}{rgb}{0.56,0.692,0.195}
\definecolor{ourviolet}{rgb}{0.528,0.471,0.701}

\newcommand{\R}{\mathbb{R}}

\newcommand{\ub}[1]{\overline{ #1 }}
\newcommand{\lb}[1]{\underline{ #1 }}
\newcommand{\norm}[1]{\left\lVert #1 \right\rVert}

\newcommand{\epssetR}{\mathcal{G}_{\epsilon}}
\newcommand{\ie}{\emph{i.e.},{ }}

\acrodef{OCP}{optimal control problem}
\acrodef{MPC}{model predictive control}
\acrodef{NMPC}{Nonlinear model predictive control}
\acrodef{EE}{end effector}
\acrodef{PDE}{partial differential equations}
\acrodef{ODE}{ordinary differential equations}
\acrodef{FEM}{finite element method}
\acrodef{AMM}{assumed mode method}
\acrodef{LPM}{lumped parameter method}
\acrodef{SF}{predictive safety filter}
\acrodef{IL}{imitation learning}
\acrodef{NLP}{nonlinear program}
\acrodef{PTP}{point-to-point}
\acrodef{EKF}{extended Kalman filter}
\acrodef{NN}{neural networks}
\acrodef{SNN}{neural network with safety filter}
\acrodef{SF}{safety filters}
\acrodef{rfes}{rigid finite elements}
\acrodef{ABA}{articulated body algorithm}
\acrodef{RNEA}{recursive Newton-Euler algorithm}
\acrodef{IL}{imitation learning}
\acrodef{RL}{reinforcement learning}
\acrodef{DOF}{degrees of freedom}

\title{\LARGE \bf
Safe Imitation Learning of Nonlinear Model Predictive Control\\ for Flexible Robots
}

\author{Shamil Mamedov$^{1,2}$, Rudolf Reiter$^{3}$, Seyed Mahdi B. Azad$^{4}$, Ruan  Viljoen$^{1,2}$, \\ Joschka Boedecker$^{4,6}$, Moritz Diehl$^{3,5}$, Jan Swevers$^{1,2}$ 
\thanks{$^1$The MECO Research Team, KU Leuven, Leuven, Belgium. }%
\thanks{$^2$The DMMS Lab, Flanders Make, Leuven, Belgium.}%
\thanks{$^3$Department of Microsystems Engineering, University of Freiburg, Freiburg, Germany.}
\thanks{$^4$Department of Computer Science, University of Freiburg, Freiburg, Germany.}
\thanks{$^5$Department of Mathematics, University of Freiburg, Freiburg, Germany.}
\thanks{$^6$BrainLinks-BrainTools, University of Freiburg, Freiburg, Germany.}
}

\makeatletter
\let\NAT@parse\undefined
\makeatother

\usepackage[hidelinks]{hyperref}
\hypersetup{
	colorlinks=true,       
	linkcolor=ourblue,        
	citecolor=ourviolet,        
	filecolor=ourblue,     
	urlcolor=ourviolet
}

\begin{document}

\maketitle
\thispagestyle{empty}
\pagestyle{empty}

\begin{abstract}
Flexible robots may overcome some of the industry's major challenges, such as enabling intrinsically safe human-robot collaboration and achieving a higher payload-to-mass ratio. However, controlling flexible robots is complicated due to their complex dynamics, which include oscillatory behavior and a high-dimensional state space. \ac{NMPC} offers an effective means to control such robots, but its significant computational demand often limits its application in real-time scenarios. To enable fast control of flexible robots, we propose a framework for a safe approximation of \ac{NMPC} using imitation learning and a predictive safety filter. Our framework significantly reduces computation time while incurring a slight loss in performance. Compared to \ac{NMPC}, our framework shows more than an \emph{eightfold} improvement in computation time when controlling a three-dimensional flexible robot arm in simulation, all while guaranteeing safety constraints. Notably, our approach outperforms state-of-the-art reinforcement learning methods.
The development of fast and safe approximate \ac{NMPC} holds the potential to accelerate the adoption of flexible robots in industry. 
The project code is available at: \href{https://tinyurl.com/anmpc4fr}{tinyurl.com/anmpc4fr}
\end{abstract}

\section{Introduction}
In recent years, flexible robots have drawn increasing attention as they may hold the key to solving significant problems in industry. These problems include improving the payload-to-mass ratio, as well as making robots intrinsically safer to facilitate human-robot collaboration \cite{DeLuca2008FlexRobots}. The primary reason why flexible robots have not yet been widely adopted is their inherent flexibility, which causes oscillations and static deflections, complicating modeling and control.

Modeling flexible robots is challenging because they possess an infinite number of \ac{DOF} and are governed by nonlinear partial differential equations. Spatial discretization converts these equations into ordinary differential equations, rendering them more suitable for control and trajectory planning. Although various modeling methods exist for flexible robots \cite{book1984recursive, shabana2020dynamics, Yoshikawa1996}, this paper adopts a lumped parameter approach \cite{Yoshikawa1996, franke2009vibration, staufer2012ella, moberg2014modeling}, which balances computational efficiency and accuracy. Having a model lays the groundwork for leveraging model-based methods, particularly optimal control, to ensure safety and high performance. An alternative approach is model-free control, particularly model-free reinforcement learning (RL). However, model-free approaches often suffer from sample inefficiency \cite{model_free_sample_eff}.

This paper specifically focuses on using model predictive control (MPC) for the output regulation problem of flexible robots: the task of reaching a desired end-effector position from an arbitrary initial configuration. The inherent nonlinearity of flexible robot dynamics necessitates the adoption of nonlinear MPC (NMPC). However, deploying NMPC for real-time control of flexible robots is often impossible due to slow computation time caused by the high-dimensional state space and nonlinear dynamics. Prior attempts to mitigate this issue have leveraged linearized MPC approaches, yielding overly conservative control strategies. In contrast, we propose an alternative approach that approximates NMPC using a neural network (or any suitable function approximator), a technique known as imitation learning. A major drawback of the approximated policy is the lack of safety guarantees. We address this by incorporating a safety filter \cite{Wabersich2021PredictiveSafetyFilter} that modifies the candidate action proposed by the learned controller if it results in a constraint violation.

To summarize, this paper leverages ideas from optimal control, imitation learning, and safety-critical control, proposing a framework for the accurate and safe output regulation of flexible robots. This framework represents the main contribution of this paper and has been validated through extensive simulation experiments. 
Specifically, we demonstrate that our framework not only significantly reduces the computational time of control action but also effectively filters out unsafe control actions, closely approaching the performance of NMPC. 
Moreover, it outperforms state-of-the-art RL algorithms. 

\section{Related work}
We organize the related work around key components of the proposed framework. 

\subsection{Modeling flexible robots}
Spatial discretization serves as the primary tool for deriving computationally tractable models of flexible robots. There are three main methods for discretizing a flexible link: the \ac{AMM} \cite{book1984recursive, green2004LQR2link}, the \ac{FEM} \cite{sunada1981application, shabana2020dynamics} and the \ac{LPM} \cite{Yoshikawa1996, franke2009vibration, staufer2012ella, moberg2014modeling}. All methods generally assume small deformations and employ the linear theory of elasticity. The \ac{FEM} is the most accurate among the three but results in a higher number of differential equations.  The \ac{AMM} is often utilized for one \ac{DOF} flexible robots; for robots with higher \ac{DOF}, the choice of boundary conditions becomes nontrivial \cite{heckmann2010choice}. The \ac{LPM} is the simplest among all, but tuning the parameters of such models may require significant effort. In this paper, we leverage the \ac{LPM} following a formulation defined in \cite{wittbrodt2007FREM} because of its simplicity and ability to exploit existing efficient rigid-body dynamics algorithms. 

\subsection{Optimal control of flexible robots}
Numerous control methods have been proposed for controlling flexible robots, including optimal control methods. Green et al.  \cite{green2004LQR2link} applied a linear quadratic controller for trajectory tracking control of a two-link flexible robot, utilizing linearized dynamics. Both \cite{silva2020implementable} and  \cite{boscariol2010model} used MPC to control a single-link flexible robot and four-link flexible mechanisms, respectively. In these papers, the authors linearized the dynamics and used linear MPC, highlighting the difficulty in developing fast \ac{NMPC} for robots with high-dimensional dynamics. Contrary to the approaches taken by \cite{silva2020implementable, boscariol2010model}, we first formulate an NMPC for the output regulation of flexible robots that fully leverages nonlinear dynamics and then approximate it.

\subsection{Approximating model predictive controllers}
To broaden the application of computationally demanding \ac{NMPC} to fast robotics systems, recent efforts have aimed at approximating \ac{NMPC} with \ac{NN}. 
Both \cite{Nubert2020} and \cite{ANMPC_pHRI} approximated robust and nominal \ac{NMPC} for motion planning in fully-actuated rigid robot arms, respectively. Nurbayeva et al. \cite{ANMPC_pHRI} did not explicitly address the safety of the approximated policy, while Nubert et al. \cite{Nubert2020} applied a statistical validation technique to ensure the safety of the approximated \ac{NMPC}, despite its time-consuming nature. Carius et al.  \cite{carius_2020} proposed a policy search method guided by \ac{NMPC} without safety considerations.

Approximating NMPC with a NN falls within the broader domain of \ac{IL}. A straightforward \ac{IL} method known as behavioral cloning (BC) \cite{Pomerleau-1989-15721} approximates NMPC (the expert policy) by addressing a supervised regression problem on a dataset  $\mathcal{D}$ containing state-action pairs. However, BC is limited as it can occasionally make mistakes, causing it to venture into areas of the state space not covered in $\mathcal{D}$, resulting in unpredictable behavior and poor performance. 
The DAgger algorithm \cite{ross2011dagger} overcomes BC's shortcomings by collecting additional expert demonstrations~$\mathcal{D}_\mathrm{new}$ as the robot interacts with the environment under the state distribution induced by the learned policy. Essentially, DAgger iteratively learns from the expert to mitigate the learned policy's approximation errors.
In contrast to BC and DAgger, Inverse Reinforcement Learning (IRL) aims to infer the expert's reward function from the dataset $\mathcal{D}$ and train a policy based on this learned reward function. Although IRL can yield more robust policies, these methods are often difficult to train due to their adversarial nature \cite{Roth2017StabilizingTO}.

Given that the expert NMPC controller can be readily queried by the DAgger during training, we incorporate DAgger into the proposed framework, unlike \cite{Nubert2020,carius_2020,ANMPC_pHRI} who employed BC. Our ablation studies (Section \ref{subsec: IL_ablation}) demonstrate DAgger's superiority over other imitation learning algorithms.

\subsection{Guaranteeing safety of learned controllers}
While approximation reduces the computational demands of deploying  NMPC, it does not inherently guarantee safety.
Recently, several methods have been proposed to ensure the safety of learned black-box controllers \cite{brunke2022safe}.  Control barrier functions \cite{CBF} and \ac{SF} \cite{Wabersich2021PredictiveSafetyFilter} stand out as two prominent approaches. The former requires an explicit definition of the safe set and operates reactively, while the latter implicitly defines the safe set and acts predictively. Given the challenge of explicitly specifying safe sets for flexible robots, this paper utilizes a \ac{SF}: an \ac{NMPC} scheme that evaluates a candidate action provided by the learned controller to verify that it can drive the system to a safe terminal set. If so, the action is directly applied to the system; otherwise, it is modified as little as possible to ensure safety.


\section{Problem statement}
\begin{figure}[t]
    \captionsetup{font=small}
    \centering
    \includegraphics[width=0.85\linewidth, trim={0cm 1.0cm 0cm 0}, clip]{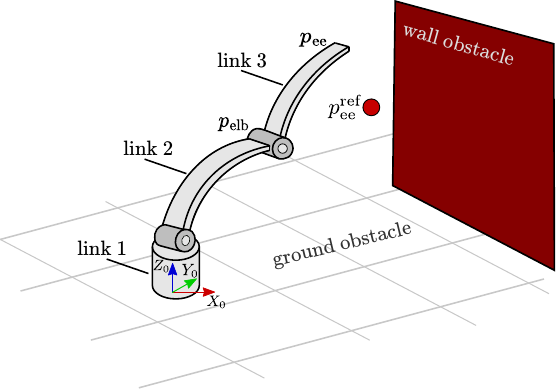}
    \caption{An illustration of the problem addressed in this paper. The controller has to move the flexible robot's end-effector $p_\mathrm{ee}$ to the goal position $p_\mathrm{ee}^\mathrm{ref}$ while guaranteeing safety in terms of preventing the robot from colliding with the \textit{wall} and \textit{ground} obstacles. }
    \label{fig:prob_stmnt}
    \vspace{-5mm}
\end{figure}

Consider the three \ac{DOF} flexible robot arms shown in Fig.~\ref{fig:prob_stmnt}, inspired by the flexible robots TUDOR \cite{malzahn2010tudor} and ELLA \cite{staufer2012ella}. The first link is rigid, while the second and third links are flexible, sharing identical dimensions and material properties. The joints are assumed to be rigid and directly actuated, meaning that the actuators do not include gearboxes.
Let $\bm x_\mathrm{tr} \in \R^{n_{x_\mathrm{tr}}}$, $\bm u \in \R^{3}$ and $ \bm z := \bm p_\mathrm{ee} \in \R^{3}$ denote the \emph{true} state, control inputs, and controlled output (end-effector position) of a flexible robot, respectively.  Furthermore, let the dynamics and output equations be defined in state-space form as follows: 
\begin{align}
    \dot{\bm x}_\mathrm{tr} = \bm f_\mathrm{tr}(\bm x_\mathrm{tr}, \bm u); \quad \bm z = \bm h_\mathrm{tr}(\bm x_\mathrm{tr}). \label{eq:generic_model}
\end{align}
where $\bm f_\mathrm{tr}(\bm x_\mathrm{tr}, \bm u)$ and $\bm h_\mathrm{tr}(\bm x_\mathrm{tr})$ represent \emph{true} nonlinear dynamics and controlled output maps, respectively.
\begin{figure*}
    \captionsetup{font=small}
    \centering
    \includegraphics[width=0.90\textwidth]{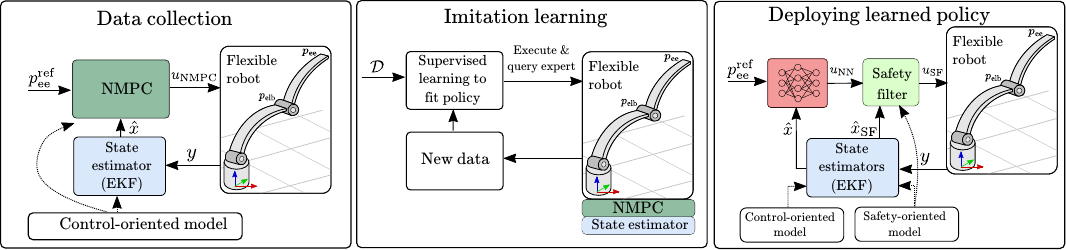}
    \caption{Proposed framework for safely approximating NMPC policy by combining imitation learning of NMPC and a safety filter.}
    \label{fig:SANMPC_pipeline}
    \vspace{-6mm}
\end{figure*}
The problem addressed in this paper is accurate and real-time feasible output regulation: steering the robot from an initial rest state $\bm x_{\mathrm{tr},0}$ and corresponding output $\bm z = \bm h_\mathrm{tr}( \bm x_{\mathrm{tr},0})$ to a final output $\bm z^\mathrm{ref} := \bm p_\mathrm{ee}^\mathrm{ref}$. The controller must adhere to the robot's constraints, such as velocity and input/torque limits while avoiding obstacles, including a \textit{wall} and \textit{ground} obstacles, as illustrated in Fig. \ref{fig:prob_stmnt}.
The measurements $\bm y = [\bm q_\mathrm{a}^\top\ \bm{\dot q}_\mathrm{a}^\top\ \bm p_\mathrm{ee}^\top ]^\top \in \R^{9}$ include the positions $\bm q_a \in \R^{3}$ and velocities $\bm{\dot q}_a \in \R^{3}$ of the actuated joints, as well as the position of the \ac{EE} $\bm p_\mathrm{ee}$, to monitor elastic deformations.

\section{Method} \label{sec:method}
\ac{NMPC} can effectively address the constrained output regulation problem for flexible robots by formulating an \ac{OCP} as a \ac{NLP} and leveraging nonlinear optimization solvers to find a sequence of control inputs that steers the robot towards the desired output \cite{Rawlings2017}. This \ac{NLP} is then repeatedly solved online for a receding horizon. However, due to the complexity of a flexible robot's dynamics and its high-dimensional state space, solving the optimization problem becomes too slow for real-time control.

To mitigate this issue and make \ac{NMPC} available for flexible robots, we propose a framework, as illustrated in Fig. \ref{fig:SANMPC_pipeline}, that safely approximates \ac{NMPC}. The approach involves three steps. First, it formulates \ac{NMPC} for the flexible robot, denoted by $\bm{\pi}_\text{NMPC}$, and generates a dataset $\mathcal{D}$ by deploying $\bm{\pi}_\text{NMPC}$ in simulation. Second, it employs the imitation learning algorithm to approximate $\bm{\pi}_\text{NMPC}$ using $\mathcal{D}$, along with an additional dataset $\mathcal{D}_\mathrm{new}$ collected during interaction with the robot and the expert \ac{NMPC}, thereby yielding $\bm \pi_{\mathrm{NN}}$. Third, it deploys $\bm \pi_{\mathrm{NN}}$ using a safety filter to ensure safety and prevent constraint violations. 

The simulation, control-oriented, and safety-oriented models differ in state space dimensions because of the different spatial discretizations employed. Specifically, the simulation utilizes a higher fidelity model with finer spatial discretization. Consequently, an \ac{EKF} is used to estimate the states of the control- and safety-oriented models based on measurements obtained from the simulation.
The following subsections describe all the components of the framework.

\subsection{Modeling and simulation of flexible robots}
\begin{figure}[t]
    \captionsetup{font=small}
    \centering
    \includegraphics[width=0.75\linewidth]{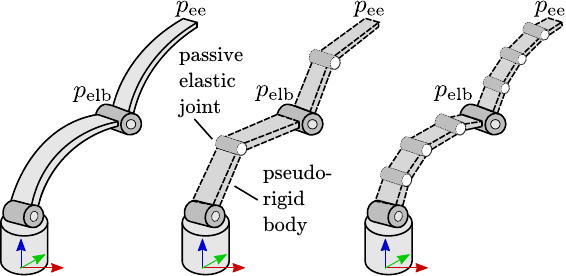}
    \caption{Illustration of two different levels of spatial discretization: one-segment (middle) and three-segment (right) discretizations.}
    \vspace{-6mm}
    \label{fig:model_discr}
\end{figure}

To model the robot, we employ an LPM, specifically the pseudo-rigid body method, which approximates a flexible link with a finite number of rigid links connected by elastic joints \cite{wittbrodt2007FREM, moberg2014modeling}, as illustrated in Fig. \ref{fig:model_discr}. This spatial discretization process involves two main steps. First, a flexible link is divided into $n_{\mathrm{seg}}$ segments, with their spring and damping properties lumped at a single point (primary division). Then, pseudo-rigid links are isolated between massless passive elastic joints (secondary division). 

%
This spatial discretization enables the application of algorithms from rigid-body literature for approximating the kinematics and dynamics of the flexible robot \cite{sciavicco2001book}. 
We denote the vector of joint angles $\bm q = [\bm q_\mathrm{a}^\top\ \ \bm q_\mathrm{p}^\top]^\top$, with $\bm q_\mathrm{p} \in \R^{2 n_{\mathrm{seg}}}$  representing passive elastic joint angles, typically modeled as spherical joints to capture the compliance in all directions. 
However, when compliance is primarily in one direction, as in our setup, simplifying the model to focus on this direction is beneficial.
Our model specifically considers lateral bending, as shown in Fig. \ref{fig:model_discr}.

Applying the Lagrange method yields the final expression for the forward dynamics of the flexible manipulator
\begin{align}
     \Ddot{\bm q} = \underbrace{\bm M( \bm q)^{-1}\big\{\bm B \bm \tau - \bm C(\bm q, \Dot{\bm q}) \Dot{\bm q} - \bm K \bm q - \bm D \Dot{\bm q} - \bm g(\bm q)\big\}}_{\mathrm{fwd}(\bm q, \dot{\bm q})} \label{eq: rgb_dyn}
\end{align}
where $n_{\mathrm{rb}} = 3 + 2 n_{\mathrm{seg}}$;  $\bm M \in \R^{n_{\mathrm{rb}} \times n_{\mathrm{rb}}}$ is the symmetric inertia matrix;   $\bm K, \bm D \in \R^{n_{\mathrm{rb}} \times n_{\mathrm{rb}}}$ are the constant diagonal stiffness and damping matrices, respectively; $\bm C \in \R^{n_{\mathrm{rb}} \times n_{\mathrm{rb}}}$ is the matrix of centrifugal and Coriolis forces, $\bm g \in \R^{n_{\mathrm{rb}}}$ is the vector of gravitational forces,  $\bm B \in \R^{n_{\mathrm{rb}} \times 3}$ is the constant control jacobian and $\bm \tau \in \R^{3}$ is the torque vector. The model is converted to state-space form by
defining   $\bm x := \left[\bm q^\top\ \dot{\bm q}^\top \right]^\top$ and $\bm u := \bm \tau$
\begin{align}
    \dot{\bm x} &= \bm f( \bm x, \bm u; n_{\mathrm{seg}}) = \left[\dot{\bm q}^\top\  \mathrm{fwd}(\bm q, \dot{\bm q})^\top\right]^\top. \label{eq:ss_dyn}  
\end{align} 
Similarly, the forward kinematics of a serial chain provides expressions for computing the positions of the end-effector~$\bm p_{\mathrm{ee}}$ and the elbow of the robot:
\begin{align}
    \bm p_\mathrm{ee} = \bm h(\bm x), \quad \bm p_\mathrm{elb} = \bm \kappa(\bm x).
\end{align}

Note that the model's accuracy and computational complexity depend on the level of spatial discretization~$n_{\mathrm{seg}}$. Therefore, we consider the model with $n_{\mathrm{seg}} = 10$ as the ground truth dynamics \eqref{eq:generic_model} used for simulation (further increasing $n_{\mathrm{seg}}$ does not improve simulation accuracy). For the control-oriented and safety-oriented models used by the NMPC and safety filter, respectively, we employed coarser but computationally faster models. Material properties of the flexible links are presented in Table \ref{tab:mater_prop}.

\begin{table}[]
    \captionsetup{font=small}
    \centering
    \begin{tabular}{cccccc}
        \toprule
        $L$ [cm] & $a$ [cm] & $h$ [cm] & $\rho$ [kg/m${}^3$] & $E$ [GPa] & $\eta \times 10^9$  \\
        \midrule
        $50$ & $5$ & $0.2$ & $7870$ & $190$ & $0.85$\\
         \bottomrule
    \end{tabular}
    \caption{Material properties of the flexible links. $L$, $a$, $h$ are length, width, and height, respectively. $\rho$ is the density, $E$ is the Young's modulus and $\eta$ is the normal damping coefficient.}
    \label{tab:mater_prop}
    \vspace{-6mm}
\end{table}

For the efficient implementation of \eqref{eq:ss_dyn}, we use the articulated body algorithm for forward dynamics and the forward path of the recursive Newton-Euler algorithm for forward kinematics, both generated by Pinocchio \cite{carpentier2019pinocchio} as CasADi \cite{Andersson2019casadi} functions. To simulate the flexible robot, we employ an implicit integrator (the backward differentiation formula, as implemented in CVODES \cite{hindmarsh2005sundials}), due to the stiffness of  \eqref{eq:ss_dyn}. Additionally, to simulate sensor noise, we add Gaussian noise to all measurements~$\bm{y}$.
Considering variations in $n_{\mathrm{seg}}$ between the control and the simulation models, and the presence of measurement noise, we implemented an \ac{EKF} as state estimator to infer control-oriented and safety-oriented models states from the measurements~$\bm y$.

\subsection{\ac{NMPC} for output regulation of flexible robots}
\label{sec:nmpc}
In robotics, an \ac{OCP} is initially formulated in continuous time, after which it is discretized and represented as an \ac{NLP}~\cite{Rawlings2017}, which is solved at each iteration of \ac{NMPC}. Various techniques are available for translating \ac{OCP} into an NLP, and this paper adopts the direct multiple shooting approach~\cite{bock1984multiple}, where the decision variables are both states and control actions. 
A general NLP for output regulation of a flexible robot can be expressed as follows:
\begin{mini!}[3]
	{\bm X,\bm U,\bm Z,\bm \Sigma}
	{L(\bm X,\bm U,\bm Z,\bm \Sigma)}			
	{\label{eq:final_mpc_policy}} 
	{} 
	\addConstraint{\bm x_0}{= \hat{\bm x},\qquad \bm \Sigma+\Delta \geq 0,}{\ \bm x_N\in S^t} \label{eq:slack_bounds}
	\addConstraint{\bm 0}{= \bm \Phi(\bm x_{k+1},\bm x_k,\bm u_k),}{\quad k=0,\ldots,N-1} \label{eq:ocp_integration}
    \addConstraint{\bm z}{= \bm h(\bm x_k),}{\quad k=0,\ldots,N} 
	\addConstraint{\lb{\bm x}-\sigma_{x,k}}{\leq \bm x_k \leq \ub{\bm x}+\sigma_{x,k},}{\quad k=0,\ldots,N}
	\label{eq:final_mpc_policy_x}
	\addConstraint{\lb{\bm u}}{\leq \bm u_k \leq \ub{\bm u},}{\quad k=0,\ldots,N-1}
	\label{eq:final_mpc_policy_u}
        \addConstraint{\bm z}{\in\mathcal{F}(\sigma_\mathrm{ee}),}{\quad k=0,\ldots,N} \label{eq:cnstr_ee}
        \addConstraint{\bm \kappa(\bm x_k)}{\in\mathcal{F}(\sigma_\mathrm{elb}),}{\quad k=0,\ldots,N}. \label{eq:cnstr_elbow}
\end{mini!}
In this formulation, $N$ represents the prediction horizon, the decision variables $\bm X = [\bm x_0, \ldots,\bm x_{N}] \in \R^{n_x \times (N+1)}$ and $\bm U = [\bm u_0, \ldots, \bm u_{N-1}]\in\R^{n_u \times N}$ correspond to states and control inputs, respectively. Algebraic decision variables $\bm Z=\left[\bm z_0,\ldots,\bm z_{N}\right]\in\R^{n_z \times N}$ represent the controlled output of the robot.

The constraints include: the discretized nonlinear dynamics of the robot \eqref{eq:ocp_integration} as implicit integration condition; upper $\ub{\bm u}$ and lower $\lb{\bm u}$ bounds on control actions (torques) \eqref{eq:final_mpc_policy_u};  upper~$\ub{\bm x}$ and lower~$\lb{\bm x}$ bounds on states \eqref{eq:final_mpc_policy_x}, softened via slack variables~$\sigma_{x,k} \in \R^+$; safety-critical, obstacle avoidance constraints for the \ac{EE} and elbow positions, \eqref{eq:cnstr_ee} and \eqref{eq:cnstr_elbow}, respectively. 
The feasible set~$\mathcal{F}(\sigma)$ in \eqref{eq:cnstr_ee} and \eqref{eq:cnstr_elbow} is formulated as a convex set that results from the intersection of two half-spaces:
\begin{align}
	\begin{split}
	\mathcal{F}(\sigma) = 
	\left\{\bm p\in\R^3
	\middle\vert
	\begin{array}{@{}r@{}}
		\bm w_\mathrm{wall}^\top(\bm p - \bm b_\mathrm{wall}) \geq -\sigma \\
		\bm w_\mathrm{ground}^\top(\bm p - \bm b_\mathrm{ground}) \geq -\sigma
	\end{array}
	\right\},
	\end{split}
    \label{eq:feas_set}
\end{align}
where~$\sigma \in \R^+$ is a slack parameter used for numerical robustness in optimization algorithms and which is zero in the optimal solution, $\bm w_{i}$ and $\bm b_i$ are the hyperplane parameters, with $i \in \{\mathrm{wall}, \mathrm{ground} \}$.
All the slack decision variables are represented by $\bm \Sigma=[\bm \sigma_0,\ldots,\bm \sigma_N]^\top\in\R^{3\times N}$, with $\bm \sigma_k=[\sigma_{x,k}\ \sigma_{\mathrm{ee},k}\ \sigma_{\mathrm{elb},k}]^\top\in \R^3$. 

To mitigate constraint violations due to noisy measurements and model mismatch, we performed a simple heuristic-based constraint tightening via slack variables.
Specifically, we established safety margins~$\bm  \delta_x \in \R^{n_x}$ for the state and safety-critical constraints~$\bm  \delta_\mathrm{ee}$ and~$\bm  \delta_\mathrm{elb}$.
The tightening of the state, output, and elbow constraints was formulated by setting bounds on the corresponding slack variables~$\sigma_x$, $\sigma_\mathrm{ee}$ and $\sigma_\mathrm{elb}$, respectively. To concisely formulate the slack bounds \eqref{eq:slack_bounds}, we utilized the matrix~$\Delta=[\delta_x \ \delta_\mathrm{ee} \; \delta_\mathrm{elb}] \otimes \bm 1_{1\times N}$, which repeats the vector~$[\delta_x \; \delta_\mathrm{ee} \; \delta_\mathrm{elb}]$ across~$N$ times.

The objective function~$L$ penalizes the deviation from the reference controlled output position~$\bm z^\mathrm{ref}$, and, as a form of regularization, it also penalizes deviations from the reference state~$\bm x^\mathrm{ref}$ and reference torque~$\bm u^\mathrm{ref}$ using an~$\ell_2$ norm: 
\begin{align}
    &L(\bm X,\bm U,\bm Z,\bm \Sigma)=
\sum_{k=0}^{N-1} 
\norm{\bm x_k-\bm x^\mathrm{ref}}_{\bm Q}^2 + 
\norm{\bm u_k-\bm u^\mathrm{ref}}_{\bm R}^2+ \nonumber \\
&\norm{\bm z_k-\bm z^\mathrm{ref}}_{\bm P}^2 + 
\norm{\bm \sigma_k}_{\bm S}^2 + 
\norm{\bm \sigma_k}_{1,s} + \norm{\bm x_N-\bm x^\mathrm{ref}_{N}}_{\bm Q_N}^2 + \nonumber  \\
&\norm{\bm z_N-\bm z^\mathrm{ref}}_{\bm P_N}^2 +
\norm{\bm \sigma_N}_{\bm S}^2 + 
\norm{\bm \sigma_N}_{1,s}, \label{eq:objective}
\end{align}
with $\bm Q=\mathrm{diag}(\bm q)\in\R^{n_x \times n_x}$, $\bm R=\mathrm{diag}(\bm r)\in\R^{n_u \times n_u}$ and $\bm P=\mathrm{diag}(\bm p)\in\R^{n_z \times n_z}$ being stage cost weighting matrices; and $\bm Q_N\in\R^{n_x \times n_x}$ and $\bm P_N\in\R^{n_z \times n_z}$ being terminal cost weighting matrices. Additionally, the slack variables $\bm \Sigma$ are  penalized in the cost function by $\ell_1$- and squared $\ell_2$-norms by weights $\bm s\in\R^3$ and $\bm S\in \R^{3\times3}$, respectively. The latter is used to facilitate convergence.

Since the optimization problem was formulated for a finite horizon, a simple safe terminal set~$S^t = \{\bm x \in \R^{n_x} | \bm H \bm x=0\}$ was used to guarantee recursive feasibility. $S^t$
corresponds to zero joint velocities and the matrix~$\bm H$ selects the angular velocity states~$\dot{\bm q}=\bm H \bm x$ of the state vector~$\bm x$.

For solving \eqref{eq:final_mpc_policy}, we used the real-time \ac{NMPC} solver~\texttt{acados}~\cite{Verschueren2021} with the ~\texttt{HPIPM} QP solver~\cite{Frison2020}.   Hyperplane parameters for the wall are defined by $\bm w_\mathrm{wall} = [0\ 1\ 0]^\top$ and $\bm b_\mathrm{wall}=[0\ -0.15\ 0.5]^\top$ and for the ground by $\bm w_\mathrm{ground} = [0\ 0\ 1]^\top$ and $\bm b_\mathrm{ground}=[0\ 0\ 0]^\top$, respectively. The remaining parameters of the \ac{NMPC} are provided in the project's repository. 

\subsection{Imitation learning of the \ac{NMPC}}
The aim of \ac{IL} within the proposed framework is to approximate the solution of \eqref{eq:final_mpc_policy}, also referred as the expert policy, by a \ac{NN} using a dataset of expert demonstrations~$\mathcal{D} = \{(\bm u_{\mathrm{NMPC}, i}, \bm{\hat x_i}, \bm s_i) \}_{i=1}^{N}$, where $\bm u_{\mathrm{NMPC}, i}$ is the control action of \ac{NMPC}, $\bm{\hat x_i}$ is a state estimate and $\bm s_i$ represents a context. The context provides additional information about the system that is not included in state-action pairs. It includes the current and desired positions of the end-effector~$\bm p_\mathrm{ee}$ and $\bm p_\mathrm{ee}^\mathrm{ref}$, respectively, as well as the hyperplane representation of the safe-critical constraints~$\bm w_{i}$ and $\bm b_i$ with $i \in \{\mathrm{wall}, \mathrm{ground} \}$. In addition to $\mathcal{D}$, DAgger collects a dataset~$\mathcal{D}_\mathrm{new}$ during interaction with the flexible robot and the expert \ac{NMPC}. 

We leveraged the ~\texttt{imitation}~\cite{gleave2022imitation} implementation of DAgger, which started from a dataset~$\mathcal{D}$ of 100 expert demonstrations and collected 500 more demonstrations during training. Key hyperparameters employed during training are provided in the project repository. 

\subsection{Safety filter}
The safety filter can be interpreted as an implicit formulation of the safe set, which is closely related to the formulation of an \ac{NMPC}. 
However, since the safety filter is used only for projecting the \ac{NN} control actions into a safe set and not for performance, the model it uses is simpler and faster to integrate, and the prediction horizon can be shorter.
The proposed safety filter policy~${\bm \pi_\mathrm{SF}:\R^{n_u}\times\R^{n_x}\rightarrow \R^{n_u}}$ projects the NN output~$\bm u_\mathrm{NN}\in \R^{n_u}$ to a safe control~$\bm u_\mathrm{SF}=\bm \pi_\mathrm{SF}(\bm x_\mathrm{SF},\bm u_\mathrm{NN}) \in\mathcal{U}_\mathrm{SF}\subseteq\R^{n_u}$ by using the same formulation as in~\eqref{eq:final_mpc_policy}, but with modified cost~$L_\mathrm{SF}$, a lower fidelity model~$\bm x_{\mathrm{SF}, {i+1}}=\bm  \Phi_\mathrm{SF}(\bm x_{\mathrm{SF}, {i}},\bm u_{\mathrm{SF}, {i}})$ with just one segment ($n_\mathrm{seg} = 1$), as shown in Fig. \ref{fig:model_discr}, and a shorter horizon. The safety filter cost function 
\begin{align}
L_\mathrm{SF}(\bm X_\mathrm{SF},\bm U_\mathrm{SF},\bm \Sigma_\mathrm{SF})=
\norm{\bm u_0-\bm u_\mathrm{NN}}_{\bar{\bm R}}^2  + L_\mathrm{REG}
\label{eq:objective}
\end{align}
penalizes the deviation from the proposed control~$\bm u_\mathrm{NN}$,  besides a regularization term
\begin{align}
\begin{split}
    &L_\mathrm{REG}(\bm X_\mathrm{SF},\bm U_\mathrm{SF},\bm \Sigma_\mathrm{SF}) = \sum_{k=1}^{N_\mathrm{SF}-1} \norm{\bm u_k}_{\bm R_\mathrm{SF}}^2 +\\ 
&\quad \quad \quad \sum_{k=0}^{N_\mathrm{SF}} \norm{\bm x_k-\bm x^\mathrm{ref}}_{\bm Q_\mathrm{SF}}^2 + 
\norm{\bm \sigma_k}_{\bm S_\mathrm{SF}}^2 + 
\norm{\bm \sigma_k}_{1,s_\mathrm{SF}} ,
\end{split}
\end{align}
with small weights~$\bm Q_\mathrm{SF}$, $\bm R_\mathrm{SF}$, $\bm S_\mathrm{SF}$, and used for numerical robustness. While \eqref{eq:objective} looks similar to 
\eqref{eq:final_mpc_policy}, due to a simpler model with fewer segments, it can be solved much faster.

\section{Experimental validation} \label{sec:exper}

To evaluate the proposed approach, we conducted extensive simulation experiments where the initial configuration of the robot and desired goal position of the \ac{EE} were randomly sampled. The goal positions were specifically chosen near obstacles to test safety. Additionally, we evaluated the policies' robustness against model uncertainties -- reduction in the stiffness parameters of the flexible robot by 10\%, in particular the $\bm K$ matrix in \eqref{eq: rgb_dyn}. For evaluation, we considered metrics such as the final distance to the goal at the end of each episode/rollout, policy evaluation time (inference time), and the number of constraint violations during an episode.

\subsection{Reinforcement learning baseline}
As model-free RL baselines, we trained SAC \cite{haarnoja2018soft} and PPO \cite{schulman2017proximal} as the state-of-the-art representatives of offline and online RL algorithms, respectively. Controlling the flexible robot by means of RL is not the main focus of this paper; thus, we did not engineer the reward function $r$. Moreover, reward engineering is difficult in practice and can lead to unintended consequences \cite{clark2016faulty}. Instead, we simply used the Euclidean distance between the robot's EE and the target EE positions, $r = - \|\bm p_\mathrm{ee} - \bm p_\mathrm{ee}^\mathrm{ref}\|_2$. We penalized the agent with $r_\mathrm{pen}$ if it violated the safety-critical constraints. 
Both algorithms were trained for 2M gradient steps. When tested on 100 regulation tasks, SAC considerably outperformed PPO. Therefore, in further experiments, only SAC was used as an RL baseline.

\subsection{Selecting a control-oriented model for NMPC}
To analyze the influence of the model fidelity (\ie the number of segments) on the controller performance, we compared the \ac{NMPC} for $n_\mathrm{seg}\in\{0,1,2,3,5,10\}$. 
The performance was evaluated by measuring the computation time~$t_\mathrm{NMPC}$, the time~$t_{\epssetR}$  that \ac{NMPC} needed to steer the \ac{EE} into an epsilon ball~$\norm{\bm z(t)-\bm z^\mathrm{ref}}\leq \epsilon$ around the reference position~$\bm z^\mathrm{ref}$ ($\epsilon=10^{-3}$ in our case) and the smallest ball with radius~$d$ that includes all points after 3 seconds
\begin{align}
    d_{\epssetR}=\min_{d}\norm{\bm z(t)-\bm z^\mathrm{ref}}\leq d \quad
    \quad \mathrm{s.t.} \; t\geq 3\mathrm{s}.
\end{align}
For $n_\mathrm{seg}\in \{0,1\}$ segments, the controller was unstable, and for ten segments, the computation time was unacceptably long. As shown in Tab.~\ref{tab:cntr_per}, the average mean computation time~$\bar t_{\mathrm{NMPC}}$ and the maximum computation time~$t_{\mathrm{NMPC}}^\mathrm{max}$ significantly increase with the number of segments. However, the performance measured in terms of $t_{\epssetR}$ and $d_{\epssetR}$ does not significantly improve. Therefore, we selected $n_\mathrm{seg}=3$ as a compromise between performance and computation time for further use with \ac{IL} algorithms.

\subsection{Simulation results}

\begin{table}[t]
    \captionsetup{font=small}
    \centering
    \begin{tabular}[t]{ccccc}
        \toprule
        $n_\text{seg}$ & $d_{{\epssetR}}/d_{{\epssetR}}^{*}$ & $t_{{\epssetR}}/t_{{\epssetR}}^{*}$ & $\bar t_{\mathrm{NMPC}}/ \bar t_{\mathrm{NMPC}}^{*}$ & $t_{\mathrm{NMPC}}^\mathrm{max} / t_{\mathrm{NMPC}}^{\mathrm{max},\ *}$ \\
        \midrule
        $3$ & $1.008$ & $0.955$ & $1.653$ & $1.589$ \\
        $5$ & $1.026$ & $0.974$ & $2.95$ & $2.940$ \\
        \bottomrule 
    \end{tabular}
    \caption{Performance comparison of \ac{NMPC} controllers with different models given by the number of segments $n_{\mathrm{seg}}$. $\bar t_{\mathrm{NMPC}}$ and $t_{\mathrm{NMPC}}^\mathrm{max}$ are average and maximum \ac{NMPC} computation times, respectively. $t_{\epssetR}$ is the time to reach an $\epsilon$-ball around the reference output; $d_{\epssetR}$ is the radius of a ball that includes all points after 3 seconds.  Evaluation metrics are measured relative to the \ac{NMPC} with $n_{\mathrm{seg}}=2$, denoted with superscript ${}^*$. }
    \label{tab:cntr_per}
    \vspace{-6mm}
\end{table}

\subsubsection{Closed loop performance}
\begin{figure}[]
    \captionsetup{font=small}
    \centering
    \includegraphics[width=0.93\linewidth,  trim={0 2mm 0 2mm}, clip]{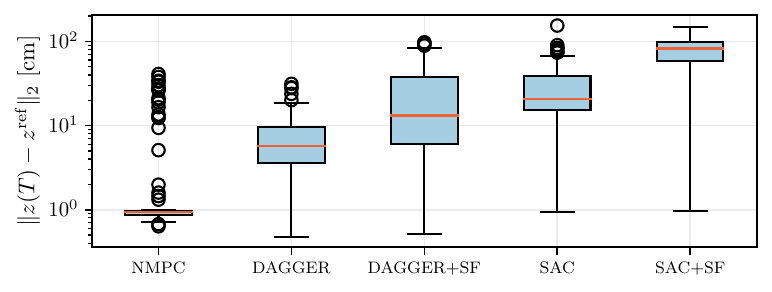}
    \caption{The distribution of final distance-to-goal ($\|z(T) - z^\mathrm{ref} \|_2$ with $T$ being the episode duration) of the considered algorithms on 100 test tasks with and without safety filter.}
    \label{fig:distances_all_w_wo_sf}
    \vspace{-4mm}
\end{figure}

Fig.~\ref{fig:distances_all_w_wo_sf} shows that \ac{NMPC} outperforms the model-free RL algorithm SAC in the output regulation of the flexible robot both in terms of accuracy measured by final distance to goal and the number of constraint violations. While DAgger (\ie \ac{IL} of NMPC) does not match the performance of the expert \ac{NMPC}, it outperforms SAC.  

Both DAgger and SAC algorithms violate safety-critical constraints, such as the wall and the ground, as shown in Fig. \ref{fig:distance_constr_time}. Importantly, both algorithms operate without explicit awareness of these constraints. RL agents learn about constraints indirectly through a scalar reward function, necessitating reward engineering to balance objectives like safety and goal achievement.  In contrast,
DAgger implicitly acquires knowledge of constraints from the trajectories of the expert. Through imitation of safe $\bm{\pi}_\text{NMPC}$, DAgger violates constraints less often than SAC. 

Incorporating a \ac{SF} significantly reduces constraint violations, albeit at the cost of decreased performance and increased computation time. With SAC, the \ac{SF} does not entirely eliminate all constraint violations, suggesting that when combined with SAC, the \ac{SF} should adopt an even more conservative approach. In the case of DAgger, introducing the \ac{SF} substantially enhances safety, making it more comparable to \ac{NMPC}.
Importantly, our proposed framework achieves an \emph{eightfold} reduction in evaluation/inference time compared to \ac{NMPC}, making the proposed approach real-time feasible, as shown in Fig.~\ref{fig:distance_constr_time}.

\subsubsection{Robustness} 
\begin{figure}
    \captionsetup{font=small}
    \centering
    \includegraphics[width=0.93\linewidth]{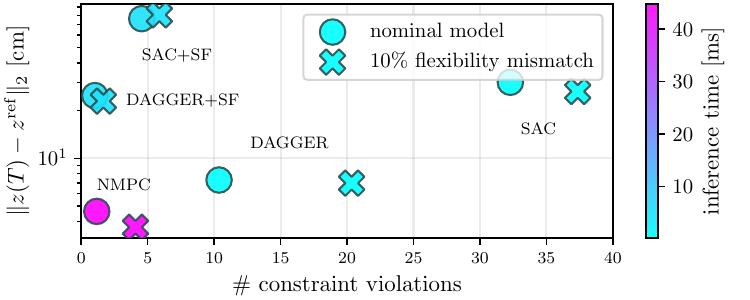}
    \caption{Performance of the considered controllers with and without model-plant mismatch. This mismatch is implemented by reducing Young's modulus of the simulation model by 10\%. The marker colors are defined by the inference time (policy evaluation time).}
    \label{fig:distance_constr_time}
    \vspace{-6mm}
\end{figure}

Notably, none of the policies were explicitly trained for robustness, except for model mismatch due to different spatial discretizations. All policies used the nominal model for generating demonstrations~$\mathcal{D}$ and for interaction.
Fig. \ref{fig:distance_constr_time} shows that changes in stiffness parameters of the simulation model affect both final distance-to-goal and constraint violation frequency. Despite the model-plant mismatch, the expert \ac{NMPC} consistently maintains high performance. In contrast, SAC and DAgger exhibit a decline in performance, particularly regarding constraint violations. The performance of the proposed framework and SAC combined with \ac{SF} remains relatively unchanged.

\subsection{Ablation of \ac{IL} algorithms} \label{subsec: IL_ablation}
To approximate NMPC, we compared BC \cite{Pomerleau-1989-15721}, DAgger \cite{ross2011dagger} and three IRL methods: GAIL \cite{ho2016generative} and AIRL \cite{fu2018learning}, and a two-step IRL method denoted as Density. GAIL and AIRL both employ adversarial approaches to jointly learn the reward function and the policy. The two-step IRL method first learns the reward function using kernel density estimation on the expert demonstrations, similar to the approach in \cite{choi2016density}, then uses the learned reward function to train a SAC agent.

\begin{figure}
    \captionsetup{font=small}
    \centering
    \includegraphics[width=0.93\linewidth, trim={0 2mm 0 2mm}, clip]{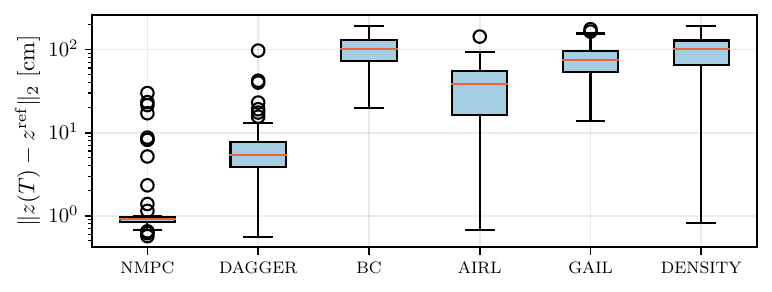}
    \caption{Comparison of different \ac{IL} algorithm for approximating \ac{NMPC} in terms of distance-to-goal at the end of the episode.}
    \label{fig:il_ablation}
    \vspace{-6mm}
\end{figure}

For training, we collected a dataset $\mathcal{D}$ of 100 expert demonstrations using \ac{NMPC} and trained all algorithms for 2 million steps. To compare \ac{IL} algorithms, we randomly generated 100 regulation tasks by sampling initial robot configurations and final end-effector positions within the robot's workspace. Figure~\ref{fig:il_ablation} shows that DAgger significantly outperforms other algorithms, making it the preferred choice for \ac{IL} of controllers like \ac{NMPC} in simulation.

\section{CONCLUSIONS}\label{sec:conc}
This paper presents a novel framework for safely approximating \ac{NMPC} for the output regulation of flexible robots. This framework integrates imitation learning with a safety filter to yield a controller that is not only computationally efficient but also accurate and safe. Initially, the framework employs DAgger for imitation learning, which approximates \ac{NMPC} using a neural network, thereby significantly reducing the computation time as compared to \ac{NMPC}. Then, a safety filter, formulated as a fast and simple \ac{NMPC}, is employed to ensure obstacle avoidance and constraint satisfaction.

This framework was validated through extensive simulations involving a three-degree-of-freedom flexible manipulator, demonstrating a significant \emph{eightfold} improvement in control action computation time. While this improvement entails some performance trade-offs, the proposed approach consistently outperformed the state-of-the-art reinforcement learning algorithm, SAC. Furthermore, we anticipate that the controller's performance will improve when more data is used to train the imitation learning component.

Additional insights gained from this research include: (i) the proposed framework demonstrates robustness to model-plant mismatches, and (ii) DAgger substantially outperforms other state-of-the-art imitation learning algorithms in approximating \ac{NMPC}. Future research could extend this approach to trajectory tracking and control challenges in soft robotics.

\addtolength{\textheight}{-3.5cm}   




\section*{ACKNOWLEDGMENT}
This work was supported by several funding agencies: FWO-Vlaanderen through SBO project ELYSA for cobot applications (S001821N); DFG via Research Unit FOR 2401 and project 424107692 and by the EU via ELO-X 953348;  the Carl Zeiss Foundation through the ReScaLe project.

\bibliographystyle{IEEEtran}
\bibliography{bibliography}

\end{document}